\documentclass[graybox]{svmult}

\usepackage{mathptmx}       
\usepackage{helvet}         
\usepackage{courier}        
\usepackage{type1cm}        
\usepackage{makeidx}         
\usepackage{graphicx}        
\usepackage{multicol}        
\usepackage[bottom]{footmisc}

\usepackage{amsmath}
\DeclareMathOperator*{\argmin}{argmin}

\makeindex             

\usepackage{cite}
\usepackage{url}
\usepackage{tikz}             
\usepackage{tikz-qtree}       
\usetikzlibrary{matrix}
\usepackage[caption=false,font=footnotesize]{subfig}
\usepackage{float}
\usepackage{algorithmic}
\usepackage{amsmath}

\usepackage{todonotes} 

\begin{document}

\title*{Correlation versus RMSE Loss Functions in Symbolic Regression Tasks}
\author{Nathan Haut,
        Wolfgang Banzhaf,
		Bill Punch       %
}

\institute{Nathan Haut \at Department of Computational Mathematics, Science and Engineering, Michigan State University, East Lansing, MI, USA  \email{hautnath@msu.edu}
\and Wolfgang Banzhaf \at Department of Computer Science and Engineering, Michigan State University, East Lansing, MI, USA \email{banzhafw@msu.edu}
\and
Bill Punch \at Department of Computational Mathematics, Science and Engineering, Michigan State University, East Lansing, MI, USA  \email{punch@msu.edu}
}

\maketitle

\abstract{The use of correlation as a fitness function is explored in symbolic regression tasks and its performance is compared against a more typical RMSE fitness function. Using correlation with an alignment step to conclude the evolution led to significant performance gains over RMSE as a fitness function. Employing correlation as a fitness function led to solutions being found in fewer generations compared to RMSE. We also found that fewer data points were needed in a training set to discover correct equations. The Feynman Symbolic Regression Benchmark as well as several other old and recent GP benchmark problems were used to evaluate performance.}


\section{Introduction}

Symbolic Regression (SR) has long been a hallmark of Genetic Programming applications. Already in John Koza's first book on GP it features prominently among the problems of program induction that he mentions as problems Genetic Programming can attempt to solve, on in the first table on page 15 \cite{koza1992}. Among the many applications listed in that table, optimal control, empirical discovery and forecasting, symbolic integration or differentiation, inverse problems and discovering mathematical identities could be easily coached in terms of symbolic regression. To cite the author: {\it ''[...], symbolic regression involves finding a model that fits a given sample of data''}. The model is constructed from mathematical functions and their numeric coefficients {\it ''that provides a good, best or perfect fit''}. 

In the absence of substantial progress on the automatic programming front in the early years of GP, symbolic regression and its Machine Learning (ML) cousin of pattern classification have taken on extremely important roles in Genetic Programming. In fact, regression and classification are now often used as {\it the} classical application examples of Genetic Programming, recognizing the contributions GP has made to those fields in the preceding decades. 

The textbook on Genetic Programming by one of the authors of this chapter \cite{bnkf1998} argued strongly in favor of the view that GP can be considered part of the Machine Learning field. To apply GP to those tasks seemed to us at the time a low-hanging fruit, and the success and widespread use of GP in these applications today proves that point. However, the second and probably more ambitious goal of GP is {\it automatic programming} in the general sense, which comprises program synthesis, program repair, probably program analysis, etc. We pointed out this ambition at the time, but did not see how to proceed until evolutionary and genetic improvement \cite{white2011,petke2017} came around. These two techniques have now thoroughly paved the way for the more general goal of GP, but it remains to be seen what results can be harvested from these techniques and re-invigorated approaches of GP to automatic programming.

Turning back to symbolic regression, Affenzeller et al \cite{affenzeller2009} explain the key task of SR as finding a [symbolic, mathematical] relationship between a dependent variable $y$ (output) and a set of specified independent (input) variables $\vec{x}$
\begin{equation}
    y = f(\vec{x},\vec{a}) + \epsilon
\end{equation}
with $f$ the functional relationship, $\vec{a}$ some coefficients to modify the functional structure, and $\epsilon$ a noise term. Here we have taken the liberty to modify the authors equation by introducing a vector notation for inputs and coefficients (note that these vectors have different dimensionality). We would also like to emphasize that the additive noise is but one type of possible noise in the system, as well as that the task could comprise the finding of relationships for more than one output variable $y$, thus rendering both $y$ and $f$ as vectors. Both input and target values are normally given as data points, perhaps produced by measurements of a system, with the expectation that the SR algorithm produces a mathematical model that is able to reproduce those data points and both interpolate among them and extrapolate beyond them.  

In recent years, a number of different non-GP symbolic regression methods have been proposed. One such method is the physics-inspired method by Udrescu and Tegmark \cite{udrescu2020} which made use of physical knowledge to restrict the search space of model creation in order to arrive at solutions in reasonable time. These authors also proposed collections of 100 new benchmark datasets (including a set of further {\it bonus} datasets) based on the known relationship of physical quantities and correspondingly derived data. Among the physical knowledge injected into the search process were considerations of unit dimensionality, translational symmetry, and multiplicative separability of the resulting models. 

What struck us was the efficiency with which certain equations could be found, sometimes even with 10 data points. At the same time we realized that the physical inspiration, while a strength in terms of knowledge injection into the process, was an ad-hoc solution that could not be transferred easily to systems of an unknown type. It seemed to us that the emphasis on global features of a model (rather than a point-by-point comparison of data points) was hinted in these results. We also knew that researchers had used other fitness functions with good outcomes, notably the Pearson correlation coefficient. 

In this chapter we report on experiments with GP symbolic regression where the fitness function, the traditional loss function called root mean square error (RMSE) 
\begin{equation}
    L= \sqrt{\frac{1}{N}\sum_{i=1}^{N} (y_i - \hat{y}_i)^2}
\end{equation}
where $N$ is the number of data points $i$, $y_i$ is the target output, and $\hat{y}_i$ the output calculated by the program under consideration,
is replaced by the correlation function
\begin{equation}
\label{eq:corr}
    R = \frac{\sum_{i=1}^N (y_i - \bar{y}) (\hat{y}_i - \bar{\hat{y}})}{\sqrt{\sum_{i=1}^N (y_i - \bar{y})^2 \times \sum_{i=1}^N (\hat{y}_i - \bar{\hat{y}})^2}}
\end{equation}
of target vs. program output.

This replacement is, however not direct. First off, we try to maximize $R^2$ (or minimize $1 - R^2$), but then, in a post-processing step, we align the resulting relationship via a simple linear regression step, minimizing 
\begin{equation}
    \argmin_{a_0, a_1} \sum_{i=1}^N (|y_i - (a_1 \hat{y_i} + a_0)|)
\end{equation}

The essential difference between these two fitness functions is the {\it global} consideration the subtracted averages of equation (\ref{eq:corr}) bring in. You can note that they are in relation to their respective output data series (target or program). They are thus entering information about the {\it shape} of the entire curve into the fitness function while the absolute scaling and translation is left to the linear regression post-processing step. We could say that the correlation function looks at the relative position of data points in the target dataset, and compares that to the relative position in the program/model produced dataset.

Correlation functions have been used in the past in genetic programming and in other data analysis applications, but often for sorting out the independence and therefore relevance of input features. Thus, it was used to identify dependent features from the input and therefore fight the curse of dimensionality, certainly a legitimate application. But applying correlation as a fitness function to compare target and program output yielded results on the AIFeynman benchmark datasets that were surprising to us. In fact, when we submitted some results to GECCO 2022, reviewers were incredulous and rejected the manuscript as a full paper. Frequently, the results showed that 3 data points are sufficient to come up with the correct equation or relationship between data points.

The book chapter presented here will examine the performance of the correlation fitness function in more detail and compare it to the RMSE loss function along a number of axes - number of data points required, noise level, and dimensionality of the input.

\section{Related Work}

In \cite{livadiotis2013} the authors develop a new fitting method relying on the maximization of the correlation coefficient between two sets of data, that could well be random, or systematically related. 
The correlation coefficient between two random variables $X$ and $Y$
\begin{equation}
    R = \frac{cov(X,Y)}{\sigma_X \sigma_Y}
\end{equation}
where $X,Y$ take on values $X_1, ..., X_N$ and $Y_1, ..., Y_N$
is a well defined quantity confined to the range [-1,+1]. If $R$ is close to either 1 or -1, the data are very strongly correlated to each other, if $R$ is close to 0, there is virtually no correlation between the two series. By minimizing 
\begin{equation}
     1 - R^2
\end{equation}
the goal of this optimization is the same as the more familiar mean square error minimization. Using a continuous example (functions instead of data points, and intergrals instead of sums) the authors can derive that the correlation-based fitting method is orders of magnitude less sensitive than the MSE method. In other words, finding a fit is much easier using the correlation-based method. 

Expressed differently, by searching for the maximum correlation between two curves/sets of points, we allow for an infinite set of possible solutions (irrespective of translation and scaling), while by searching for the minimum of MSE, we allow only one solution to be possible. More recent work by one of the same authors \cite{livadiotis2020} generalizes their sensitivity analysis to an entire family of fitting methods that are based on different $L_q$ norms. 

Keijzer \cite{keijzer2003}, revisited by \cite{nicolau2020}, lays out an example of a symbolic regression problem where the issue of (R)MSE becomes clearly visible. A simple target function
\begin{equation}
    y = x^2
\end{equation}
and a modified target function 
\begin{equation}
    y = x^2 + 100 
\end{equation}
are compared in terms of their performance under symbolic regression. The authors comment that the simple addition of a constant was able to mislead their GP systems in the search space, resulting in 
only 16\% of their runs (with given representation and parameters) being successful, compared to the unmodified regression problem, which was solved by 98\% of their runs. Keijzer suggests a linear scaling method, among other measures like interval arithmetic, to address such problems. While useful under special circumstances, the simple replacement by the MSE fitness function with a correlation fitness function (plus its post-processing step) discussed here perfectly solves this problem.  

\begin{figure}[h]
\caption{Correlation as a fitness function can identify successful models that would not be identified by RMSE. The red surface represents a model that was identified as good by the correlation fitness function and the green surface is the same model after has been aligned. Orange points represent the raw data, which consists of 100 points. Both the model and the aligned model have an $R^2$ value of 0.9999. The pre-aligned model has an RMSE value of 22.24 and the aligned model has an RMSE value of $2.076*10^{-14}$ }
\label{fig:alignDemo}
\includegraphics[width=12cm]{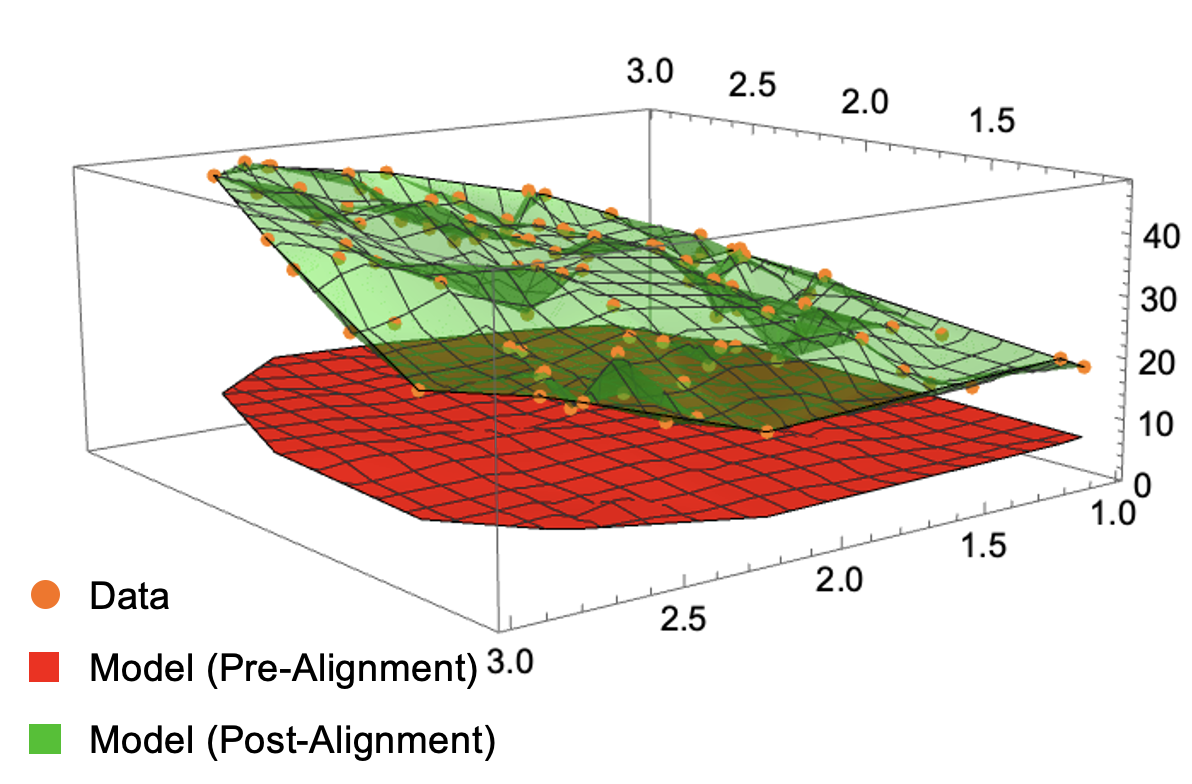}
\end{figure}

Figure \ref{fig:alignDemo} demonstrates how correlation as a fitness function is capable of identifying potentially good models that RMSE would not identify as a good model. This is a result of correlation assigning fitness independent of a linear scaling and shift, so it can identify a model that has the correct shape but may not yet be in the correct location or at the correct scale.

\section{Benchmarks}

A number of different benchmarks have been proposed for symbolic regression tasks. GP started out with the symbolic regression problems used in Koza's first book \cite{koza1992}. Over time some more complex problems were added culminating in the suggestion by White et al \cite{white2013} to consider these as new standard benchmarks. Korns \cite{korns2013,korns2014,korns2015}  has given a series of presentations at GPTP on systematically more difficult symbolic regression problems and their solution, increasingly relying on hybrid algorithms to solve them. Finally, Udrescu and Tegmark have proposed a collection of physical laws extracted from Feynman lectures \cite{udrescu2020} 
as a good benchmark suite for symbolic regression algorithms. 

In the following we shall briefly discuss the former, giving examples of each, before focusing on the latter, the so-called AIFeynman benchmark set. 

\subsection{Koza's Benchmarks}

Koza was the first to highlight the joys and sorrows of symbolic regression with Genetic Programming. The quartic polynomial 
\begin{equation}
    x^4 + x^3 + x^2 + x
\end{equation}
was the first discussed in \cite{koza1992}, though one could argue that the bang-bang control problem discussed in an earlier section was close. This problem later proliferated to the following problems:
\begin{equation}
   x^4 - x^3 + x^2 - x
\end{equation}
\begin{equation}
   x^4 + 2 x^3 + 3 x^2 + 4 x
\end{equation}
\begin{equation}
   x^6 - 2 x^4 + x^2
\end{equation}
Many more problems of different types were proposed in the book (e.g. symbolic derivatives and symbolic integration, equivalence relationships, roots of equations, etc.). 

\subsection{New Benchmark Standards}

Keijzer extended the benchmark set studied in \cite{keijzer2003} to the Keijzer instances which were further expanded by Vladislavleva et al. \cite{vladislavleva2008} and Nguyen et al. \cite{uy2011}. Typical examples are Keijzer-5:
\begin{equation}
    \frac{30 x z}{(x - 10) y^2}
\end{equation}
Vladislavleva-1:
\begin{equation}
    \frac{e^{-(x_1 - 1)^2}}{1.2 + (x_2 - 2.5)^2}
\end{equation}
or Nguyen-5:
\begin{equation}
    sin(x^2) cos(x) -1  
\end{equation}
just to name a few.

A 2012 community survey \cite{mcdermott2012}
revealed the mainly used benchmarks and was summarized and standardized in \cite{white2013}. 

\subsection{The GPTP Benchmarks}

While the term GPTP benchmarks is actually broader, we would like to focus here on the series of contributions and problem suggestions by Korns and his co-authors \cite{korns2007,korns2008large,korns2009,korns2010symbolic,korns2011abstract,korns2013,korns2014,korns2015}. 

Since this is a large set of problems, we are going to select only one here, Korns-8:
\begin{equation}\label{eq:korns8}
    6.87 + 11 \sqrt{7.23 x_0 x_3 x_4} = 6.87 + 29.58 \sqrt{x_0 x_3 x_4}
\end{equation}
out of 5 dimensions $x_0, ..., x_4$, where some variables ($x_1, x_2$) do not carry information but only noise.

\subsection{Feynman Symbolic Regression Benchmark}

Here we shall mainly focus on the AIFeynman set of equations/data, lifted out of the lectures of Richard Feynman \cite{feynman1963a,feynman1963b,feynman1963c}.

\section{Methods}

Symbolic regression was performed using StackGP, a stack-based genetic programming system. The parameters chosen for the system are shown in Table \ref{tab:parameters}. It is important to note that the two sub-populations are evolved in parallel yet do not interact until completion. Upon completion, the two populations are merged and the most fit individuals in the combined population are then selected and returned as the final population of a run. 

\begin{table}
\caption{Evolution Parameter Settings}
\label{tab:parameters}
\begin{center}
\begin{tabular}{ll} 
\hline
Parameter & Setting\\
\hline
 Mutation Rate & 79 \\ 
 Crossover Rate & 11 \\
 
 Spawn Rate & 10\\
 
 Elitism Rate & 10\\
 
 Crossover Method \hspace{5mm}  & 2-point\\
 
 Tournament Size & 5\\
 
 Population Size & 300\\
 
 Independent Runs & 100\\
 
 Sub-populations & 2\\
 
 Termination Criteria & 2 Minutes (wall time) \\
\hline
\end{tabular}
\end{center}
\end{table}

To compare the performance of using RMSE against correlation as a fitness function we explored how noise, number of points, and dimensionality affect the resulting fitnesses of the best individuals found during evolution. 

For each problem and set of conditions (noise and number of points) a total of 100 repeated independent trials were conducted and the median fitness of the best models from each trial was computed using the test data for the associated problem. To make for a simple comparison, both models trained using RMSE and correlation as their fitness functions were evaluated using RMSE on the test data. The test data consisted of 200 points generated without noise added to determine how close the evolved models are to the true generating equation. 

\subsection{Noise Introduction}

Uniformly distributed multiplicative random noise was introduced to the response data $y=f(x)$ by supplying a percentage to a noise generating function:

\begin{equation}
y = f(x) (1 + \epsilon),
\end{equation}
where 
\begin{equation}
    \epsilon \in \left[ -\frac{R}{2},\frac{R}{2} \right]
\end{equation}
is a uniformly distributed random variable from the interval $[-\frac{R}{2},\frac{R}{2}]$.

\subsection{Varying Number of Points}

For most problems, between 3 and 193 points were used to determine how changing the number of data points affects the success of the search. This was performed by initially testing with 3 points, then adding 20 points until a total of 193 points were tested. For some of the Feynman problems the number of points varied from 3 to 19 points incrementing by 2 to observe how small changes in the number of points impacts method performance. In each independent repeated trial, the points used were generated randomly anew. 

\subsection{Dimensional Sensitivity}

Sensitivity to dimensionality was explored by observing the variation in success between the different problems which vary from 1 dimension up to 9 dimensions.  

\subsection{Sensitivity to Constants}
The sensitivity to identifying equations correctly when constants are introduced was explored by introducing constants of varying magnitude and determining how the error is affected by the magnitude of constants.

\section{Results}

\subsection{The Keijzer-5 Benchmark}

\begin{figure}[h]
\caption{Comparing using RMSE and correlation as the fitness function on Keijzer-5 with 10\% noise.}
\label{fig:keijzer5}
\includegraphics[width=12cm]{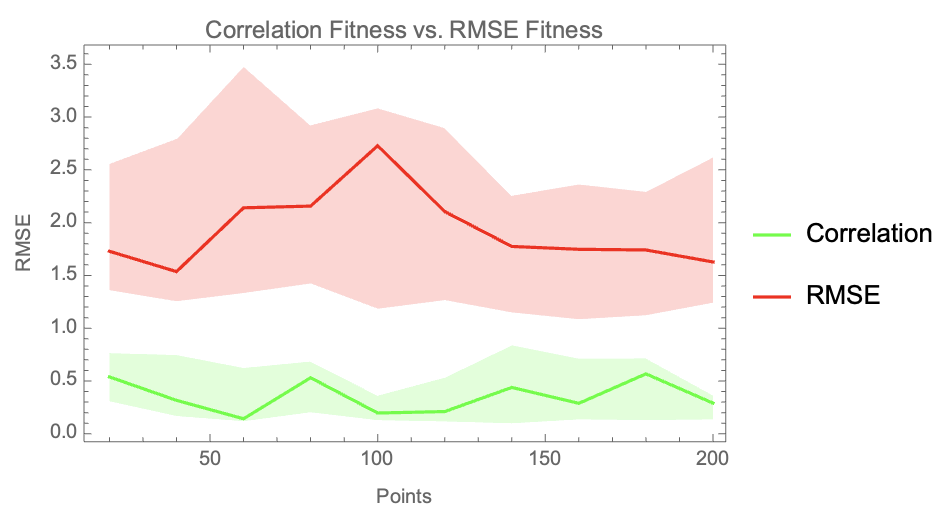}
\end{figure}

The results of comparing the correlation based fitness function to the usual RMSE on the new benchmark standards with 20 to 200 points and 10\% noise are shown in Figure \ref{fig:keijzer5}. The results show that correlation finds more accurate models than RMSE with 10\% noise for the Keijzer 5 benchmark problem. 

\begin{figure}[h]
\caption{Comparing using RMSE and correlation as the fitness function on Keijzer-5 with varying noise to determine noise tolerance when using 2,000 training points.}
\label{fig:keijzer5nt1}
\includegraphics[width=12cm]{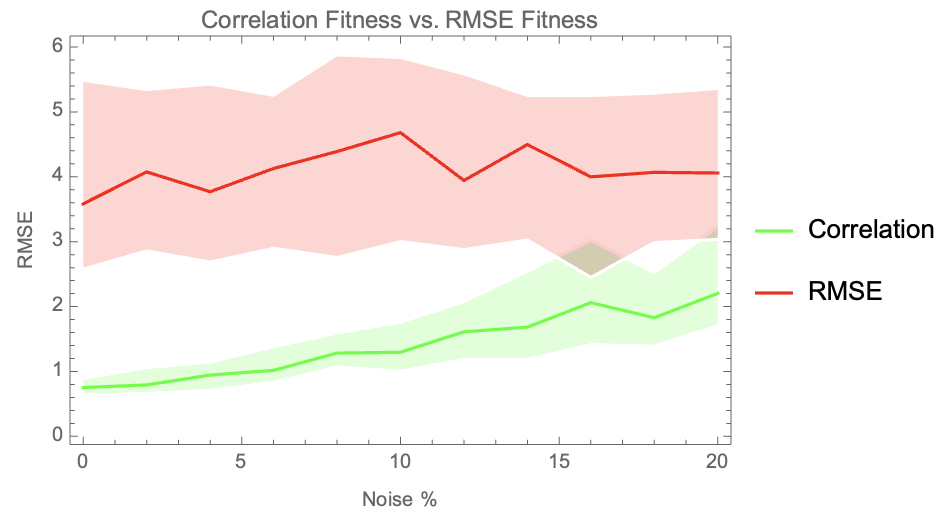}
\end{figure}

In Figure \ref{fig:keijzer5nt1} noise sensitivity was explored with 2,000 points with noise between 0 and 20\% on Keijzer-5. While the correlation approach is sensitive to multiplicative noise and gradually deteriorates as the amplitude of noise is increased, the correlation approach generally finds more accurate models even with noise as high as 20\%. 

\subsection{The Korns-8 Benchmark}

\begin{figure}[h]
\caption{Comparing using RMSE and correlation as the fitness function on Korns-8 with 0\% noise.}
\label{fig:korns8}
\includegraphics[width=12cm]{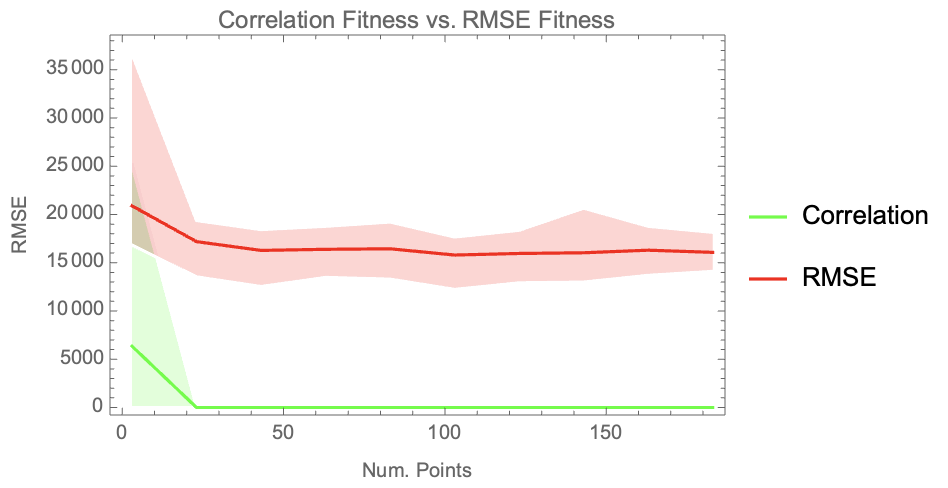}
\end{figure}

Figure \ref{fig:korns8} now compares the results of runs using correlation and RMSE as fitness functions to try to solve the Korns-8 benchmark problem, ranging from 3 to 193 training data points with no (0\%) noise.  The correlation based fitness function consistently finds essentially perfect solutions with more than 3  points, while the RMSE based fitness function performed relatively poorly for all numbers of data points, with only a slight improvement as the number of points increases.

\begin{figure}[h]
\caption{Comparing using RMSE and correlation as the fitness function on Korns-8 as noise increases from 0 to 20\% with 200 training points. With 12\% noise or more the correlation approach becomes comparable or worse than RMSE. }
\label{fig:korns8nt1}
\includegraphics[width=12cm]{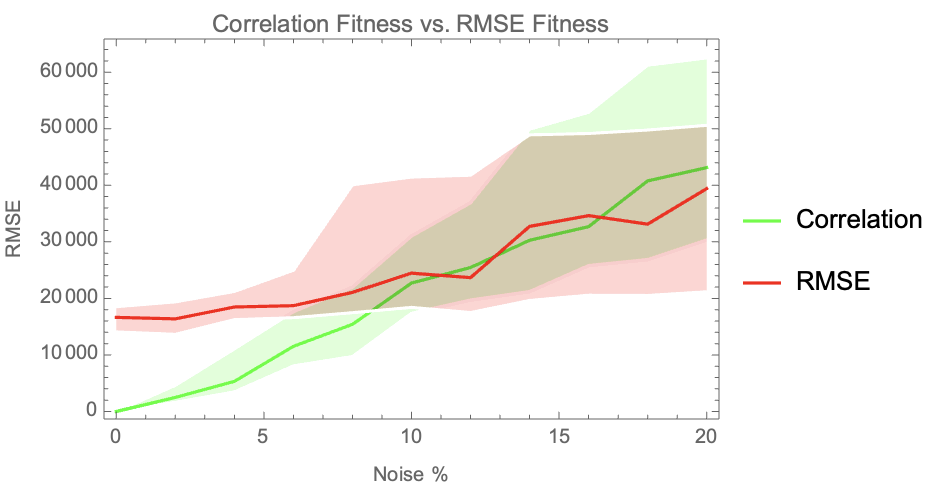}
\end{figure}

\begin{figure}[h]
\caption{Comparing using RMSE and correlation as the fitness function on Korns-8 as noise increases from 0 to 20\% with 2000 training points.}
\label{fig:korns8nt2}
\includegraphics[width=12cm]{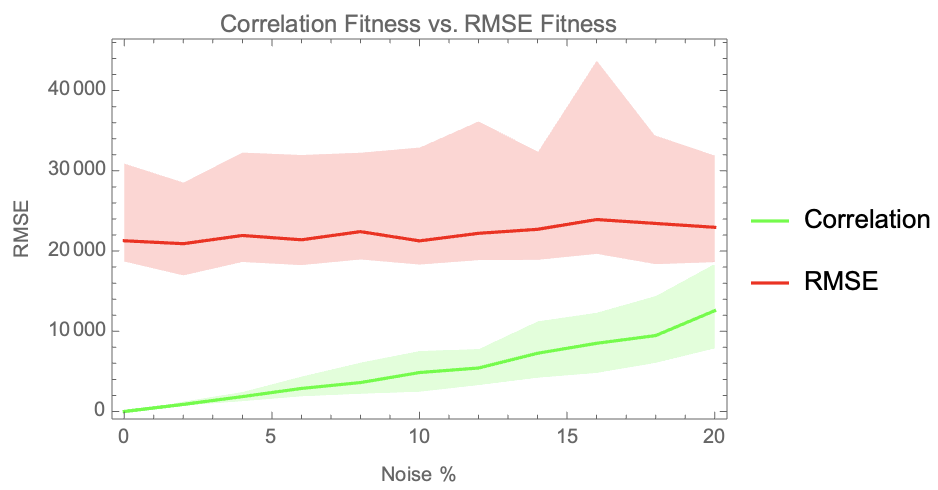}
\end{figure}

The noise tolerance of correlation and RMSE were also explored on Korns-8 by varying noise from 0\% to 20\% using 200 and 2,000 training points. The results are shown in Figures \ref{fig:korns8nt1} and \ref{fig:korns8nt2}, respectively. With 200 data points (Fig. \ref{fig:korns8nt1}) the correlation approach stops outperforming RMSE when 12\% or more noise is included in the data. When the data has 2,000 training points, however, as shown in Figure \ref{fig:korns8nt2}, we see that while the correlation approach shows sensitivity to the amount of noise present, it still finds better models with data that has 20\% noise, demonstrating that more data can effectively counter noise with a correlation fitness function.

\begin{figure}[h]
\caption{Comparing using RMSE and correlation as the fitness function on Korns-8 as noise increases from 0 to 20\% with 2,000 training points using $R^2$ as the metric for comparison.}
\label{fig:korns8nt3}
\includegraphics[width=12cm]{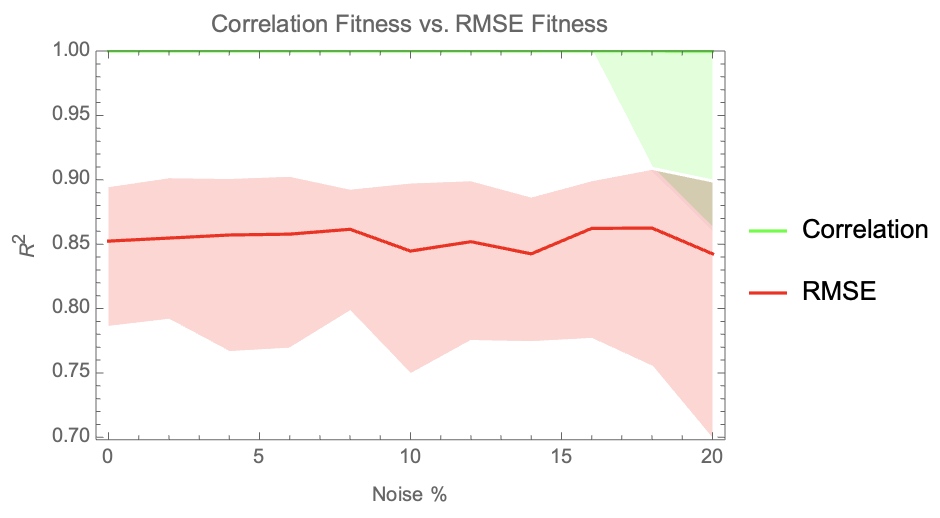}
\end{figure}

\begin{figure}[h]
\caption{Comparing using RMSE and correlation as the fitness function on Korns-8 as noise increases from 0 to 20\% with 200 training points using $R^2$ as the metric for comparison.}
\label{fig:korns8nt4}
\includegraphics[width=12cm]{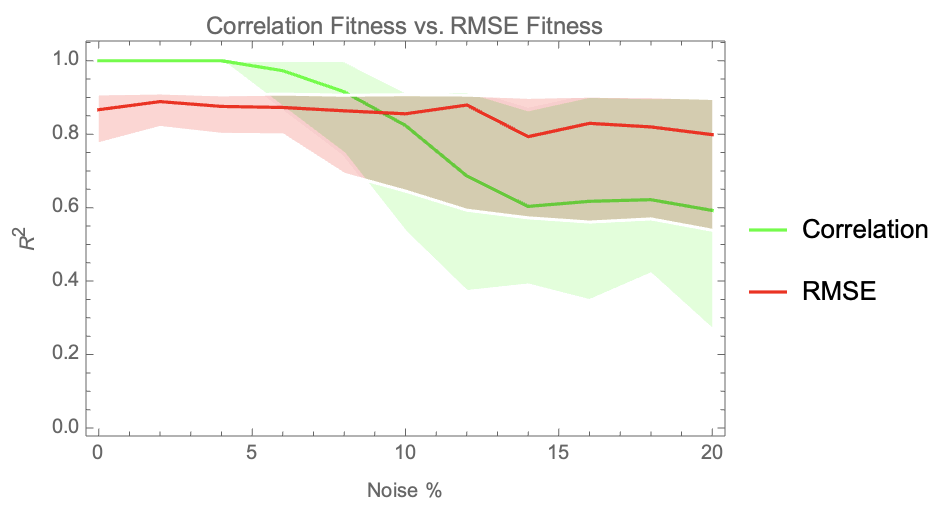}
\end{figure}

What is interesting to note is that when comparing the $R^2$ values between the two approaches with using 2,000 points, it can be seen in Figure \ref{fig:korns8nt3} that $R^2$ values for all models found using correlation are 1, up to 20\% noise. This indicates that the correct model form was still found in all cases and points to the alignment step as the part that is sensitive to noise. An example solution is shown in equation (\ref{eq:ex1}), compare to original function (eq. \ref{eq:korns8}):

\begin{equation}\label{eq:ex1}
    949.216+21.536 \sqrt{x_0 x_3 x_4}
\end{equation}
When looking at the $R^2$ values from the 200 point cases, we can see there is a very clear threshold where the signal-to-noise ratio becomes too small and the quality of models drops off significantly by using the correlation approach. The behaviour of the correlation fitness (green dots) in Fig. \ref{fig:korns8nt4} looks like a phase transition at around 10\% noise. The functional relationship evolved before the transition (e.g., at 2\% noise) is still correct:
\begin{equation}\label{eq:ex2}
    564.468+23.315 \sqrt{x_0 x_3 x_4}
\end{equation}
However, with 14\% noise (after the transition) the functional relationship is not any more correct with:
\begin{equation}\label{eq:ex3}
    1406.43+0.133 x_0 x_3 x_4
\end{equation}
The correct variables are still found, but the square root function is not any more present.

\subsection{The Vladislavleva-1 Benchmark}

\begin{figure}[h]
\caption{Comparing using RMSE and correlation as the fitness function on Vladislavleva-1 with 0\% noise as the number of points increases from 20 to 200. }
\label{fig:vlad1N0}
\includegraphics[width=12cm]{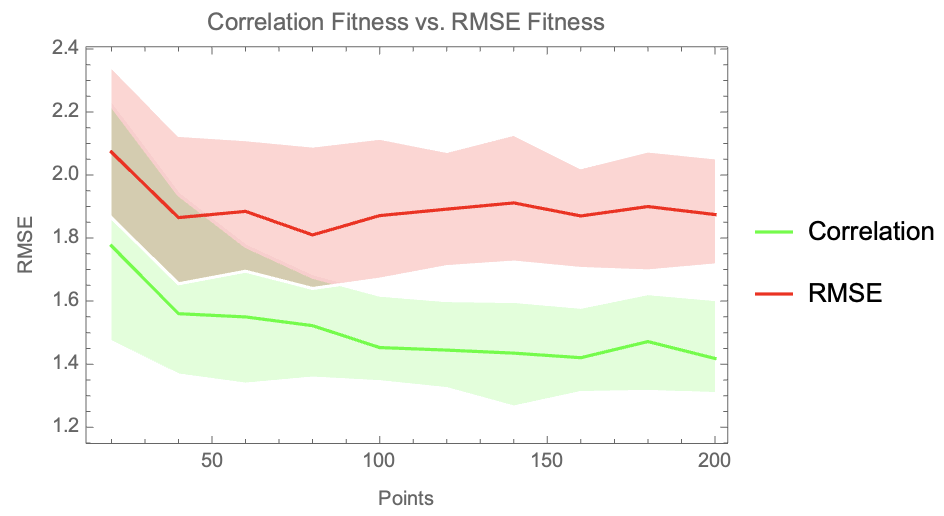}
\end{figure}

Figure \ref{fig:vlad1N0} compares the performance of using correlation and RMSE as fitness functions on the Vladislavleva-1 benchmark problem, ranging from 20 to 200 training data points with no noise. The correlation approach shows that it is able to consistently find models with better performance than when using RMSE as the fitness function. 

\begin{figure}[h]
\caption{Comparing using RMSE and correlation as the fitness function on Vladislavleva-1 as noise increases from 0 to 20\% with 200 training points using RMSE as the metric for comparison.}
\label{fig:vlad1NTol}
\includegraphics[width=12cm]{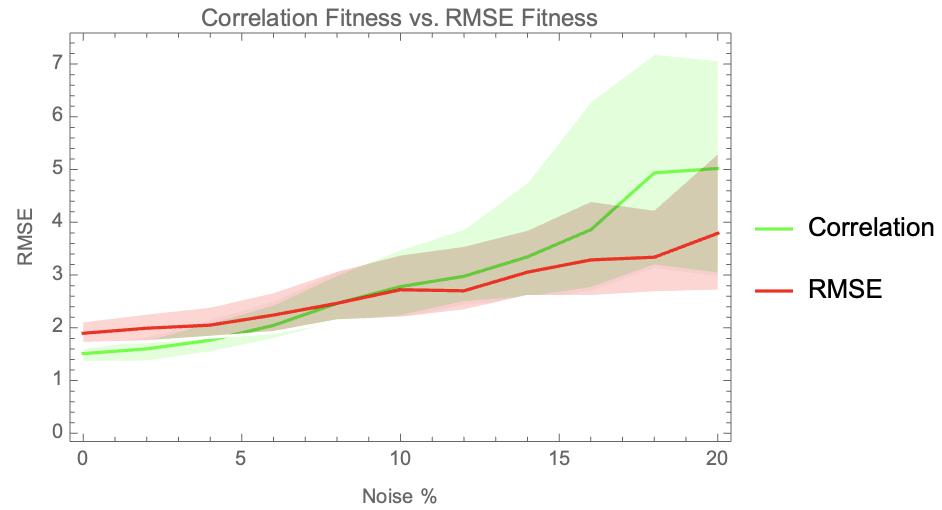}
\end{figure}

Figure \ref{fig:vlad1NTol} explores how the two approaches vary in their sensitivity to noise. Both methods were given 200 training points and noise was varied from 0 to 20\%. Correlation was observed to outperform RMSE as a fitness function until the noise level exceeded 6\%. Beyond 6\% the distributions of both methods widened significantly and the average performance of correlation as a fitness function became worse than RMSE at higher noise levels. 

\subsection{The Nguyen-5 Benchmark}

\begin{figure}[h]
\caption{Comparing using RMSE and correlation as the fitness function on Nguyen-5 with 0\% noise as the number of points varies from 20 to 200.}
\label{fig:nguyen5N0}
\includegraphics[width=12cm]{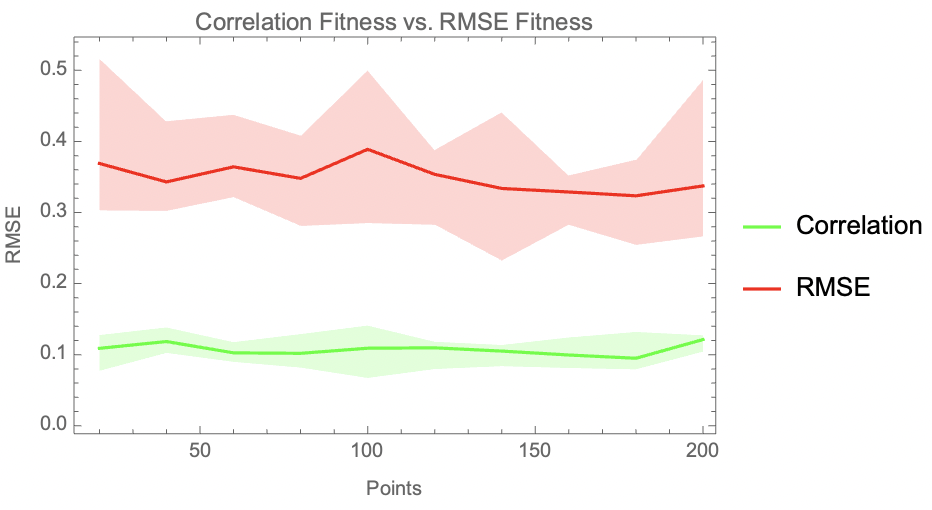}
\end{figure}

The performance of correlation and RMSE as fitness functions was also compared using the Nguyen-5 benchmark problem. Figure \ref{fig:nguyen5N0} shows how they compare when no noise is present when training on varying number of points from 20 to 200. Correlation was observed to outperform RMSE as a fitness function for all of the cases between 20 and 200 points. 

\begin{figure}[h]
\caption{Comparing using RMSE and correlation as the fitness function on Nguyen-5 as noise increases from 0 to 20\% with 200 training points using RMSE as the metric for comparison.}
\label{fig:nguyen5NTol}
\includegraphics[width=12cm]{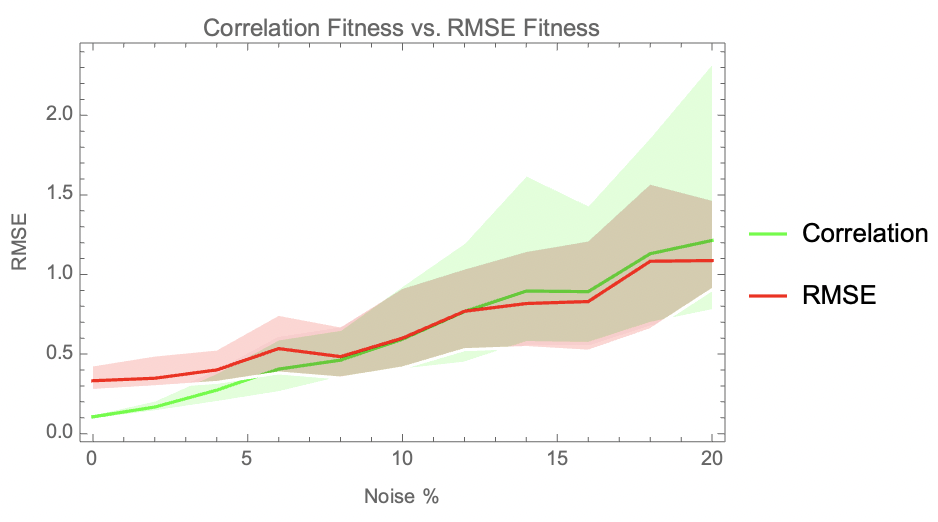}
\end{figure}

The noise tolerance of the two methods was also compared using the Nguyen-5 benchmark problem. Noise was varied from 0 to 20\% with 200 training points. The results are shown in Figure \ref{fig:nguyen5NTol}. The method using correlation as a fitness function performed best until around 6\% noise was present. Beyond 6\% noise the two methods performed similarly with correlation having a slightly wider distribution of solutions. 

\subsection{Feynman Symbolic Regression Benchmark}
The results of testing the performance of the two different fitness functions on the Feynman Symbolic Regression Benchmark are summarized in Table \ref{tab:feynman}. 

\begin{table}
\caption{Feynman Symbolic Regression Benchmark Summary Performance Comparison of Correlation Against RMSE}
\label{tab:feynman}
\begin{center}
\begin{tabular}{c|c|c|c|c|c} 
\hline
 Number of Points & Noise \% & Better & Tied & Perfectly Solved & Perfectly Solved where RMSE failed\\
\hline
 3 & 0 & 38 & 11 & 21 & 10 \\
 3 & 10 & 27 & 0 & 0 & 0 \\
 20 & 0 & 82 & 17 & 41 & 24 \\
 20 & 10 & 78 & 0 & 0 & 0\\
 200 & 0 & 81 & 17 & 46 & 29 \\
 200 & 10 & 79 & 0 & 0 & 0\\
\hline
\end{tabular}
\end{center}
\end{table}

With just 3 data points and no noise, the correlation approach found better models in 38 of the 100 cases and tied in performance with the RMSE approach in 11 cases. A total of 21 problems were perfectly solved with just 3 data points using the correlation approach and 10 of those 21 were not perfectly solved using the RMSE approach with just 3 data points. 

\begin{figure}[h]
\caption{The error of best individuals over generations is shown for equ. (\ref{eq:mu}) (\# 8 in \cite{udrescu2020}) when using correlation as the fitness function. We can see that equation (\ref{eq:mu}) is trivial and is solved almost immediately}
\label{fig:eq8conv}
\begin{center}
\includegraphics[width=8cm]{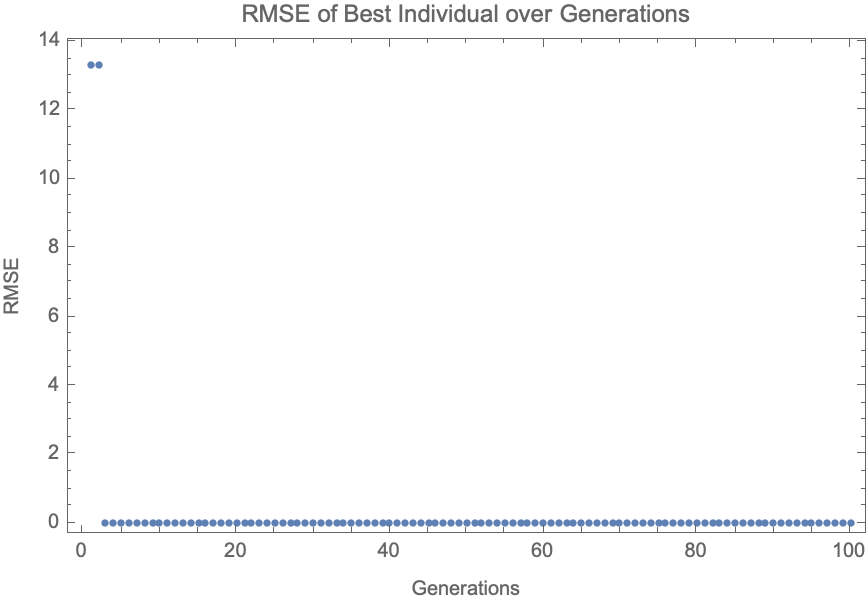}
\end{center}
\end{figure}

For example Figure \ref{fig:eq8conv} shows RMSE over generations on equ. (\ref{eq:mu}) (\# 8 in \cite{udrescu2020}) when using correlation fitness. 
\begin{equation}\label{eq:mu}
    \mu \times N_n
\end{equation}
RMSE converges to 0 very quickly in a sample evolutionary run, demonstrating that this equation is trivial to solve. 

\begin{figure}[h]
\caption{The error of best individuals over generations is shown for equ. (\ref{eq:eps}) (\# 59 in \cite{udrescu2020}) when using RMSE as the fitness function. We can see that progress seems to get stuck at a local minima and stops progressing around generation 200. }
\label{fig:eq59conv}
\begin{center}
\includegraphics[width=8cm]{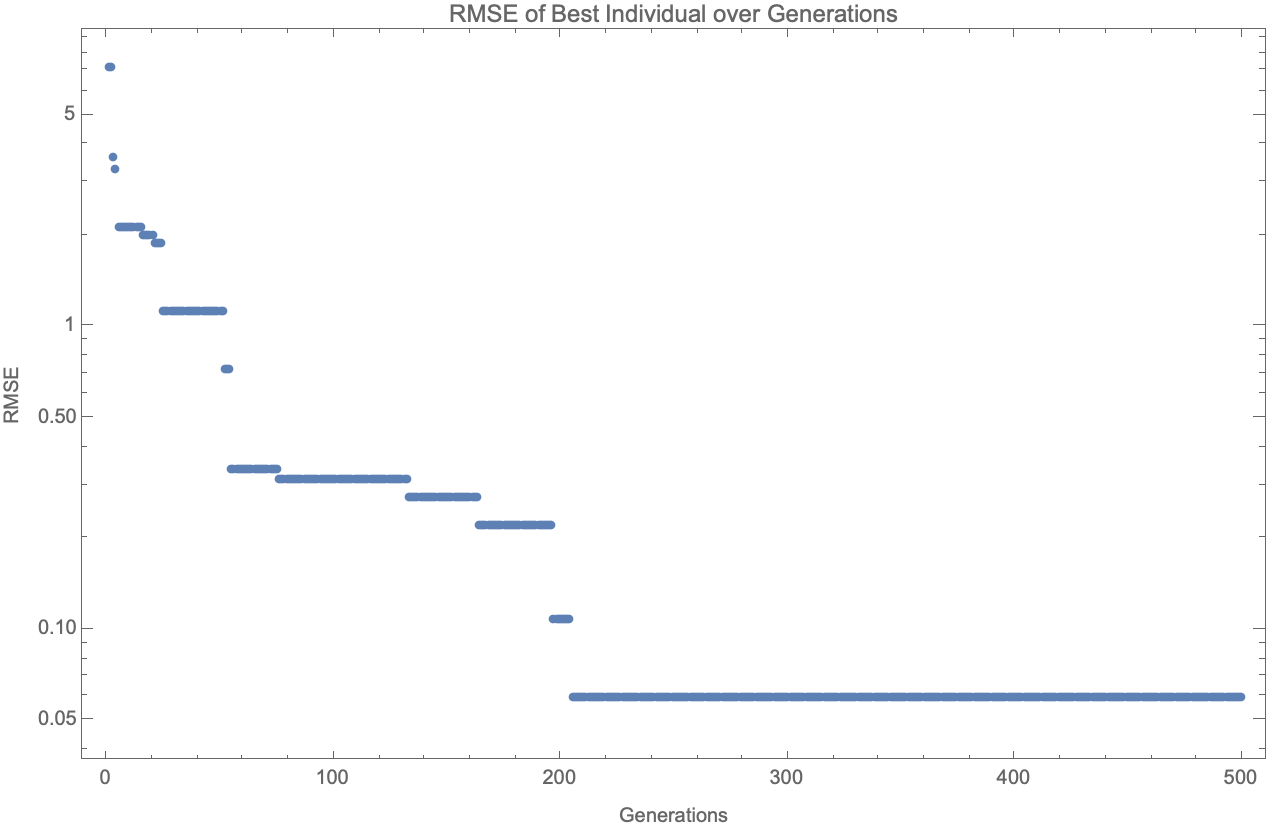}
\end{center}
\end{figure}

As another example, we take equ. (\ref{eq:eps}) (\# 59 in \cite{udrescu2020}) as one where RMSE fitness does not converge to 0, see Figure  \ref{fig:eq59conv}. 
\begin{equation}\label{eq:eps}
    \frac{\epsilon  \times E^2}{2}
\end{equation}

\begin{figure}[h]
\caption{The error of best individuals over generations is shown for equ. (\ref{eq:eps}) (\# 59 in \cite{udrescu2020}) when using correlation as the fitness function. We can see that this problem now becomes trivial and is solved by the second generation when using correlation instead of RMSE as the fitness function. }
\label{fig:eq59convcorr}
\begin{center}
\includegraphics[width=8cm]{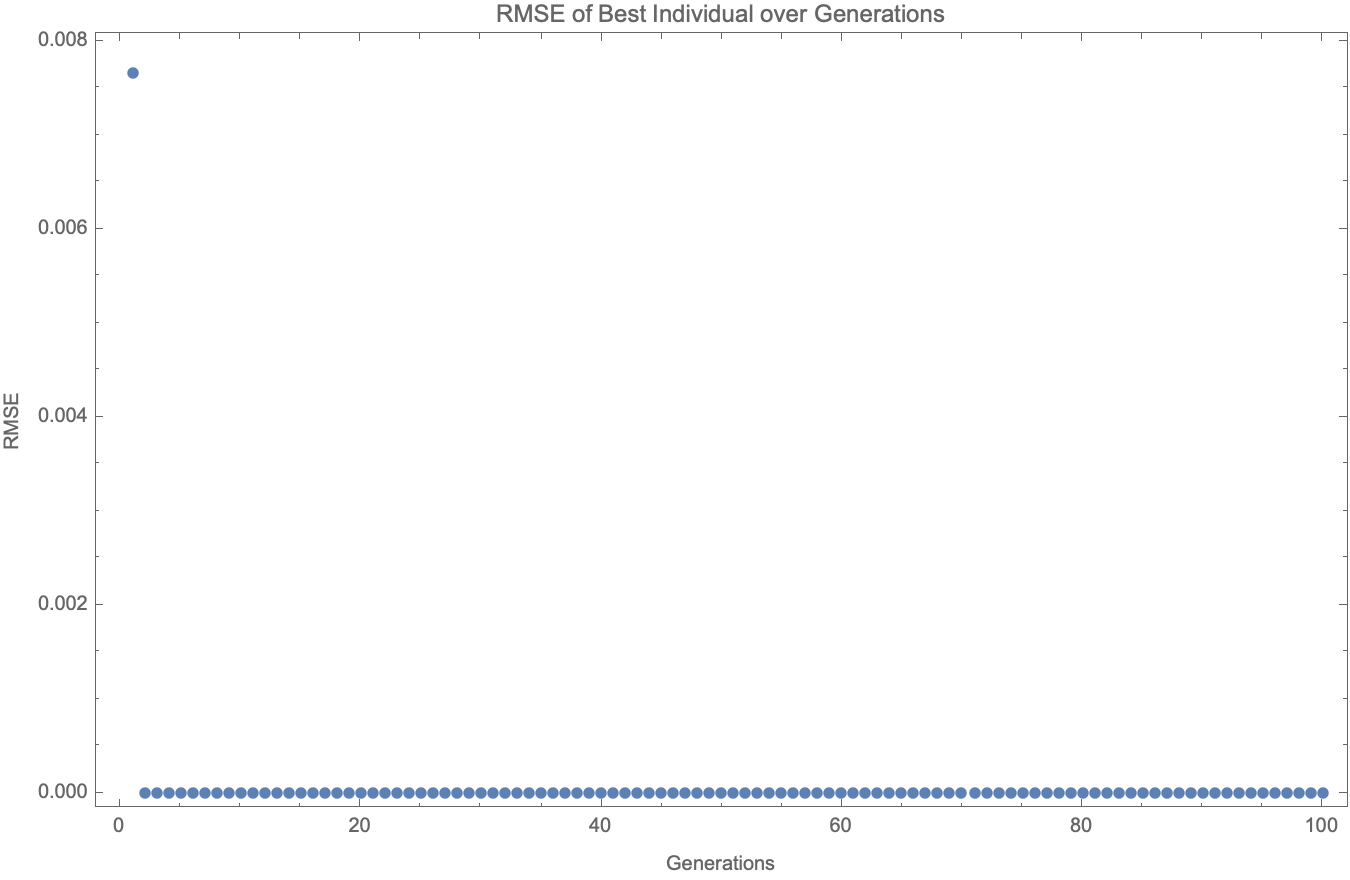}
\end{center}
\end{figure}

As opposed to that, using the correlation fitness function results in runs like that depicted in Figure \ref{fig:eq59convcorr}. As we can seem from the equations, these are very simple functional relationships. We shall examine more complicated ones later, but for now will look at noise.

With just 3 data points and 10\% noise added, the correlation approach does never converge to a perfect solution, but found better models in 27 of the 100 problems. On the other hand, RMSE is able to withstand the noise better, and is able to converge to perfect solutions 13 times.

Given that three data points really is an absolute minimum, it is no wonder that reviewers reacted incredulous to our results. But this seems more to be a reflection of the benchmark datasets, rather than the approach to solve it. 
In many cases, RMSE does also solve the problem perfectly with 3 data points (see Tables \ref{tab:pts} and \ref{tab:pts2} in the appendix).

So while on trivial cases there might not be a difference between the performance of RMSE and correlation as a fitness function, in other cases there is a difference, often a substantial one. With 20 points and no noise the correlation approach found better models in 82 of the 100 cases and was tied in another 17 cases. This means that the correlation approach beat or matched the performance of RMSE in 99 of the 100 cases of the Feynman benchmarks. As well, the correlation approach perfectly solved 41 of the problems. Of those 41, 24 were not solved using the RMSE approach. 

With 10\% noise added to the 20 data points, the correlation approach found better models in 78 of 100 cases. Here again, RMSE is better able to withstand the noise and converge to a perfect solution in 17 cases.

Finally, with 200 points and no noise, the correlation approach found better models than the RMSE approach in 81 of the 100 cases and was tied in 17 cases. This totals to 98 cases where the correlation approach either beat or matched the RMSE approach. A total of 46 cases were solved perfectly using the correlation approach. Of those 46 cases, 29 of them were not perfectly solved using the RMSE approach. 

With 200 points and 10\% noise added, the correlation approach found better models in 79 of the cases.

\begin{figure}[h]
\caption{Comparing using RMSE and correlation as the fitness function on equ. (\ref{eq:theta}) with 10\% noise.}
\label{fig:eq1n10}
\includegraphics[width=12cm]{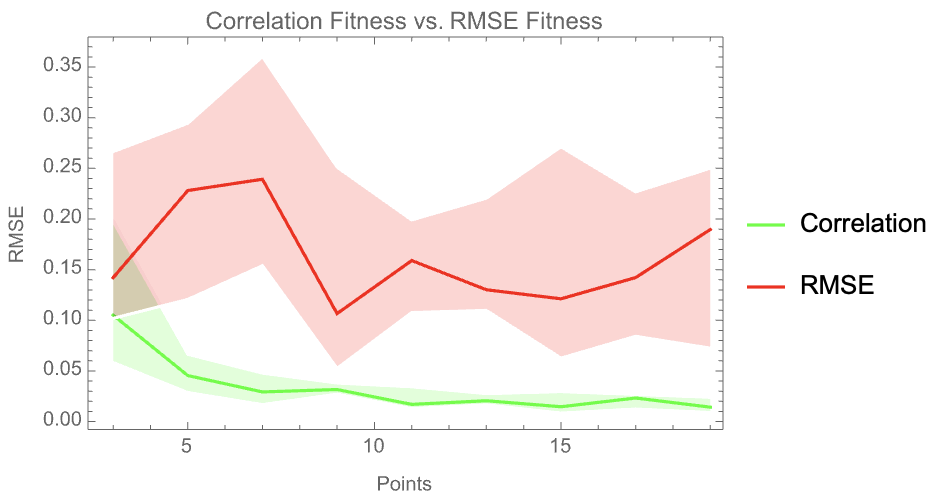}
\end{figure}

\begin{figure}[h]
\caption{Comparing using RMSE and correlation as the fitness function on equ. (\ref{eq:theta-s}) with 10\% noise.}
\label{fig:eq2n10}
\includegraphics[width=12cm]{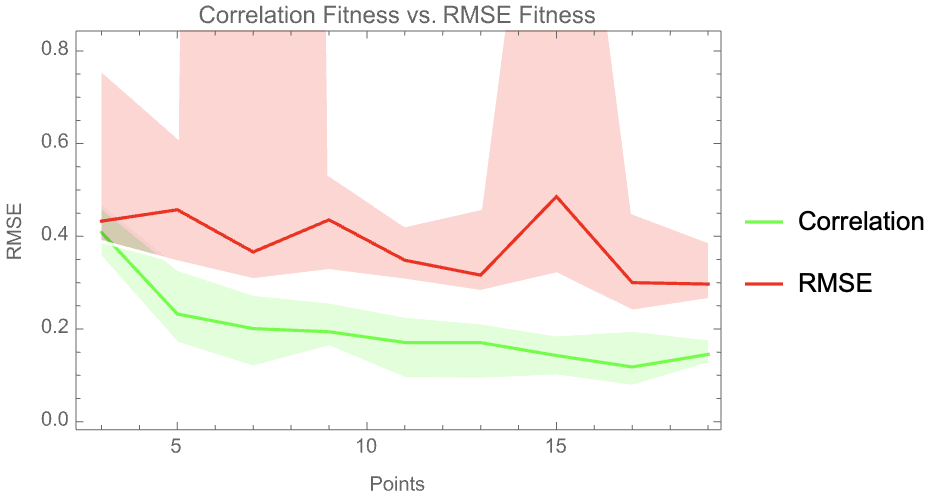}
\end{figure}

\begin{figure}[h]
\caption{Comparing using RMSE and correlation as the fitness function on equ. (\ref{eq:theta-1}) with 10\% noise.}
\label{fig:eq3n10}
\includegraphics[width=12cm]{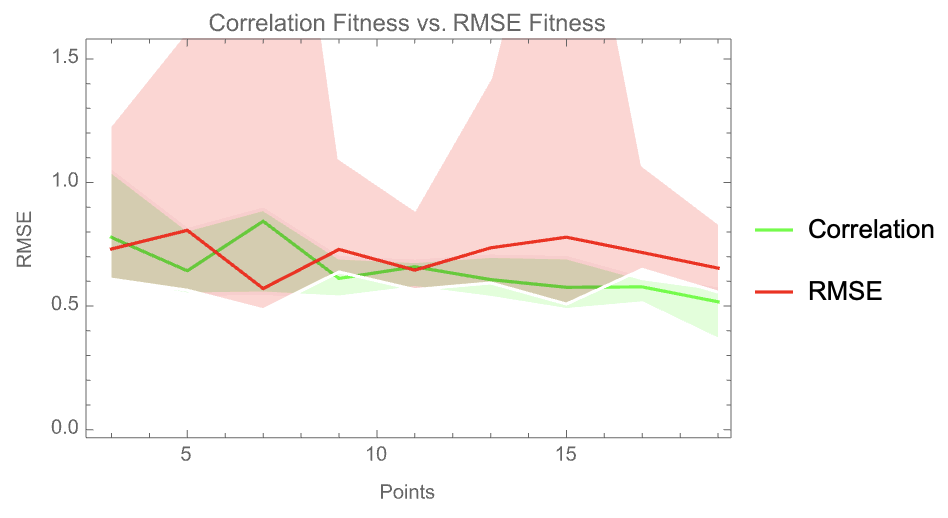}
\end{figure}

Figures \ref{fig:eq1n10}-\ref{fig:eq3n10} show three examples of noisy data and the behavior of fitness functions depending on number of data points. These are non-trivial cases not perfectly solved by either of the methods even with 200 points. However, we can clearly see the progress by correlation as opposed to RMSE.

The three equations used are
\begin{equation}\label{eq:theta}
    \frac{e^{\frac{- \theta^2}{2}}}{\sqrt{2 \pi}}
\end{equation}
\begin{equation}\label{eq:theta-s}
    \frac{e^{\frac{- (\theta/\sigma) ^2}{2}}}{\sqrt{2 \pi}\sigma}
\end{equation}
and
\begin{equation}\label{eq:theta-1}
      \frac{e^{\frac{- ((\theta - \theta_1)/\sigma) ^2}{2}}}{\sqrt{2 \pi}\sigma} 
\end{equation}

\subsection{Sensitivity to Constants}

It was observed that equation (\ref{eq:mu}) with 10\% noise was able to be found with 0 error when using RMSE as the fitness function compared to an RMSE value of 1.35 when using correlation as the fitness function when trained with 20 points. The $R^2$ values from both approaches was 1, which indicates that they both found the correct general pattern in the data, but the scaling or position was off as a result of the alignment when using correlation as fitness function. To see if the problem difficulty was altered by introduction of a constant, equation (\ref{eq:mu}) from the Feynman symbolic regression benchmark was modified by introducing a constant initially valued at 5 and then 50. The results are shown in Table \ref{tab:const8}. With no noise the problem becomes much more difficult for the RMSE fitness function approach the larger the constant, while the problem difficulty does not change relative to the constant size when using correlation as  fitness function (recall Keijzer's observation mentioned earlier). When multiplicative noise is added, the performance when using correlation does get worse as the size of the constant increases, yet it strongly outperforms when compared to using RMSE as fitness function. 
\begin{table}
\caption{Sensitivity to Constants in (\ref{eq:mu}). Feynman Equation \# 8}
\label{tab:const8}
\begin{center}
\begin{tabular}{lllll} 
\hline
Constant & Number of Points & Noise \% & RMSE & Correlation\\
\hline
 1 & 20 & 10 & 0 & 1.35 \\
 1 & 20 & 0 & 0 & 0\\
 5 & 20 & 10 & 261.69 & 8.74 \\
 5 & 20 & 0 & 276.76 & 0 \\
 50 & 20 & 10 & 3520.43 & 68.07 \\
 50 & 20 & 0 & 2151.56 & 0\\
 
\hline
\end{tabular}
\end{center}
\end{table}

\section{Discussion}

Symbolic regression has been studied since Genetic Programming was invented more than 30 years ago. Numerous refinements have been
proposed and examined, new benchmark data sets have been subjected to algorithmic variants, and results have been derived and considered in multiple forms. However, no measure has proven so efficient than simply replacing the fitness function based on absolute function values with a fitness measure that considers relative distance and leaves the absolute alignment of a functional model to a linear regression step in post-processing. Equipped with such a fitness function, many of the benchmark problems used (among them about half of the AIFeynman problems) can be safely removed from consideration as too easy. If 3 data points are sufficient to deduce the functional form of a model, then this is not a problem worthy of much attention.   

For other problems, correlation did widely outperform RMSE as a fitness function, and should be the function of first choice in all regression problems. 
It remains to be seen whether there are some other factors that allow the algorithm discussed here to shine at solving the benchmark problems considered. For example, it could be that StackGP is particularly well suited for the task and the same might not be said of other GP systems.

\section{Conclusions}
What is clear is that the fitness function in any evolutionary algorithm has an extremely important role to play as far as the performance 
of the algorithm is concerned. Lessons derived from the success of correlation-based fitness in symbolic regression might transfer to other tasks, like classification problems. At the very least, it could facilitate the solution of harder problems in other domains, by training a researcher's eye on the essential functions of a system that need to be tuned to arrive at feasible solutions.

\begin{acknowledgement}
Part of this work was funded by the Koza Endowment to MSU. Computer support by MSU's iCER high-performance
computing center is gratefully acknowledged.
\end{acknowledgement}

\footnotesize
\bibliographystyle{unsrt}
\bibliography{biblio}

\section*{Appendix}
This appendix lists all 100 AI Feynman problems and their solution using correlation and RMSE as fitness functions, for 0\% and 10\% noise levels at 3 data points only. More performance results are available online at https://tinyurl.com/stackGPGPTP.

\begin{table}
\caption{The resulting RMSE values on test data when using correlation or RMSE as the fitness function during training for first 50 Feynman equations. Three random training data points were used with either 0 or 10\% noise. }
\label{tab:pts}
\begin{center}
\begin{tabular}{cccccc}
\hline
 \text{Filename} & \text{} & \text{Correlation} & \text{RMSE} & \text{Correlation} & \text{RMSE} \\
 \text{} & \text{EQ $\#$} & \text{3 Pts 0$\%$} & \text{3 Pts 0$\%$} & \text{3 Pts 10$\%$} & \text{3 Pts 10$\%$} \\
 \hline
 \text{I.6.20a} & 1. & {\bf 0.032} & 0.304 & {\bf 0.12} & 0.35 \\
 \text{I.6.20} & 2. & {\bf 0.41} & 0.53 & {\bf 0.41} & 0.43 \\
 \text{I.6.20b} & 3. & {\bf 0.68} & 0.83 & {\bf 0.57} & 0.61 \\
 \text{I.8.14} & 4. & 21.71 & {\bf 21.61} & {\bf 11.2} & 15.66 \\
 \text{I.9.18} & 5. & {\bf 2.02} & 2.09 & {\bf 2.15} & 2.24 \\
 \text{I.10.7} & 6. & 2.29 & {\bf 1.63} & 2.74 & {\bf 1.95} \\
 \text{I.11.19} & 7. & 155.4 & {\bf 137.08} & 145.9 & {\bf 142.88} \\
 \text{I.12.1} & 8. & 0. & 0. & 3.74 & {\bf 0.} \\
 \text{I.12.2} & 9. & 291.77 & {\bf 235.69} & 302.15 & {\bf 244.97} \\
 \text{I.12.4} & 10. & 1.29 & {\bf 1.17} & 1.25 & {\bf 0.95} \\
 \text{I.12.5} & 11. & 0.29 & {\bf 0.26} & 0.29 & {\bf 0.26} \\
 \text{I.12.11} & 12. & 0. & 0. & 4.48 & {\bf 0.} \\
 \text{I.13.4} & 13. & {\bf 311.99} & 382.06 & 380.01 & {\bf 378.98} \\
 \text{I.13.12} & 14. & 123.88 & {\bf 123.} & 131.49 & {\bf 121.92} \\
 \text{I.14.3} & 15. & 0. & 0. & 16.67 & {\bf 0.} \\
 \text{I.14.4} & 16. & {\bf 0.} & 29.75 & {\bf 9.03} & 29.53 \\
 \text{I.15.3x} & 17. & 9.06 & {\bf 6.44} & 9.83 & {\bf 7.16} \\
 \text{I.15.3t} & 18. & 2.15 & {\bf 1.81} & 2.31 & {\bf 2.1} \\
 \text{I.15.1} & 19. & 4.34 & {\bf 3.57} & 4.91 & {\bf 3.54} \\
 \text{I.16.6} & 20. & {\bf 7.26} & 8.16 & {\bf 8.41} & 9.61 \\
 \text{I.18.4} & 21. & {\bf 5.2} & 6.38 & {\bf 5.73} & 6.4 \\
 \text{I.18.12} & 22. & {\bf 39.64} & 81.53 & 88.04 & {\bf 0.} \\
 \text{I.18.14} & 23. & 289.04 & {\bf 269.43} & 312.51 & {\bf 294.09} \\
 \text{I.24.6} & 24. & 186.77 & {\bf 103.88} & 172.39 & {\bf 111.05} \\
 \text{I.25.13} & 25. & 0. & 0. & 0.61 & {\bf 0.} \\
 \text{I.26.2} & 26. & 1.44 & {\bf 0.} & 4.78 & {\bf 0.} \\
 \text{I.27.6} & 27. & 4.59 & {\bf 3.53} & {\bf 3.47} & 3.82 \\
 \text{I.29.4} & 28. & 0. & 0. & 0.88 & {\bf 0.} \\
 \text{I.29.16} & 29. & 42.96 & {\bf 39.75} & 44.78 & {\bf 39.01} \\
 \text{I.30.3} & 30. & 170.3 & {\bf 47.} & 56.17 & {\bf 45.07} \\
 \text{I.30.5} & 31. & 0. & 0. & 1.07 & {\bf 0.} \\
 \text{I.32.5} & 32. & {\bf 10.48} & 11.09 & 11.02 & {\bf 10.95} \\
 \text{I.32.17} & 33. & {\bf 58.75} & 62.13 & 74.19 & {\bf 63.58} \\
 \text{I.34.8} & 34. & 118.78 & {\bf 117.31} & 152.25 & {\bf 112.47} \\
 \text{I.34.1} & 35. & 12.5 & {\bf 12.39} & 12.833 & {\bf 12.611} \\
 \text{I.34.14} & 36. & 9.69 & {\bf 6.41} & 9.27 & {\bf 8.09} \\
 \text{I.34.27} & 37. & 0. & 0. & 4.64 & {\bf 0.} \\
 \text{I.37.4} & 38. & {\bf 45.37} & 53.82 & {\bf 34.71} & 44.55 \\
 \text{I.38.12} & 39. & 809.61 & {\bf 764.83} & 797.34 & {\bf 766.62} \\
 \text{I.39.1} & 40. & {\bf 0.} & 33.53 & {\bf 6.12} & 49.26 \\
 \text{I.39.11} & 41. & 40.23 & {\bf 22.69} & 32.95 & {\bf 29.32} \\
 \text{I.39.22} & 42. & {\bf 57.45} & 109.84 & 136.6 & {\bf 116.42} \\
 \text{I.40.1} & 43. & $3.21\times 10^{12}$ & {\bf $3.12\times 10^{12}$} & $2.3\times 10^{15}$ & {\bf $9.67\times 10^{14}$} \\
 \text{I.41.16} & 44. & 31.63 & {\bf 29.5} & 31.62 & {\bf 30.15} \\
 \text{I.43.16} & 45. & {\bf 93.43} & 123.12 & 140.63 & {\bf 113.85} \\
 \text{I.43.31} & 46. & 0. & 0. & 15.21 & {\bf 0.} \\
 \text{I.43.43} & 47. & 24.15 & {\bf 20.21} & 23.18 & {\bf 19.87} \\
 \text{I.44.4} & 48. & 315.51 & {\bf 278.57} & 299.86 & {\bf 265.23} \\
 \text{I.47.23} & 49. & {\bf 5.43} & 6.32 & 7.34 & {\bf 4.88} \\
 \text{I.48.2} & 50. & 30.91 & {\bf 17.23} & 136.99 & {\bf 29.78} \\
 \hline
 \end{tabular}
\end{center}

\end{table}

\begin{table}
\caption{The resulting RMSE values on test data when using correlation or RMSE as the fitness function during training for the second set of 50 Feynman equations. Three random training data points were used with either 0 or 10\% noise.}
\label{tab:pts2}
\begin{center}
\begin{tabular}{cccccc} 
\hline
 \text{Filename} & \text{} & \text{Correlation} & \text{RMSE} & \text{Correlation} & \text{RMSE} \\
 \text{} & \text{EQ $\#$} & \text{3 Pts 0$\%$} & \text{3 Pts 0$\%$} & \text{3 Pts 10$\%$} & \text{3 Pts 10$\%$} \\
 \hline
 \text{I.50.26} & 51. & {\bf 32.39} & 32.81 & 35.81 & {\bf 33.07} \\
 \text{II.2.42} & 52. & 96.89 & {\bf 67.68} & 111.34 & {\bf 93.28} \\
 \text{II.3.24} & 53. & {\bf 0.} & 0.22 & {\bf 0.044} & 0.328 \\
 \text{II.4.23} & 54. & {\bf 0.} & 0.29 & {\bf 0.339} & 0.357 \\
 \text{II.6.11} & 55. & {\bf 0.27} & 0.28 & {\bf 0.279} & 0.362 \\
 \text{II.6.15a} & 56. & 5.59 & {\bf 4.05} & 5.01 & {\bf 4.11} \\
 \text{II.6.15b} & 57. & 40.16 & {\bf 39.29} & 29.19 & {\bf 28.66} \\
 \text{II.8.7} & 58. & {\bf 1.06} & 1.24 & {\bf 1.106} & 1.116 \\
 \text{II.8.31} & 59. & {\bf 0.} & 35.63 & {\bf 9.68} & 23.01 \\
 \text{II.10.9} & 60. & 2.97 & {\bf 1.52} & 2.74 & {\bf 1.79} \\
 \text{II.11.3} & 61. & {\bf 1.135} & 1.155 & 1.25 & {\bf 1.08} \\
 \text{II.11.17} & 62. & 27.32 & {\bf 24.32} & {\bf 20.98} & 22.98 \\
 \text{II.11.20} & 63. & 118.25 & {\bf 94.68} & 103.66 & {\bf 95.24} \\
 \text{II.11.27} & 64. & 5.24 & {\bf 2.61} & 4.21 & {\bf 2.71} \\
 \text{II.11.28} & 65. & {\bf 0.012} & 0.2 & 0.77 & {\bf 0.52} \\
 \text{II.13.17} & 66. & {\bf 0.3047} & 0.3393 & 0.36 & {\bf 0.3} \\
 \text{II.13.23} & 67. & 2.12 & {\bf 1.59} & 2.68 & {\bf 1.93} \\
 \text{II.13.34} & 68. & 3.9 & {\bf 2.76} & 5.75 & {\bf 3.76} \\
 \text{II.15.4} & 69. & {\bf 67.4} & 88.35 & 95.53 & {\bf 83.07} \\
 \text{II.15.5} & 70. & {\bf 63.28} & 80.92 & 94.52 & {\bf 92.21} \\
 \text{II.21.32} & 71. & 0.714 & {\bf 0.656} & 0.67 & {\bf 0.66} \\
 \text{II.24.17} & 72. & {\bf 1.72} & 2.31 & {\bf 2.405} & 2.963 \\
 \text{II.27.16} & 73. & 0. & 0. & 79.1 & {\bf 0.} \\
 \text{II.27.18} & 74. & 0. & 0. & 17.51 & {\bf 0.} \\
 \text{II.34.2a} & 75. & {\bf 0.} & 5.86 & {\bf 4.33} & 5.04 \\
 \text{II.34.2} & 76. & {\bf 0.} & 47.1 & {\bf 9.4} & 45. \\
 \text{II.34.11} & 77. & 65.58 & {\bf 62.45} & 73.52 & {\bf 63.65} \\
 \text{II.34.29a} & 78. & {\bf 0.} & 3.07 & {\bf 1.98} & 2.996 \\
 \text{II.34.29b} & 79. & 569.68 & {\bf 454.52} & 534.74 & {\bf 460.95} \\
 \text{II.35.18} & 80. & {\bf 4.73} & 5.1 & {\bf 4.99} & 5.36 \\
 \text{II.35.21} & 81. & 65.37 & {\bf 49.} & 58.39 & {\bf 48.6} \\
 \text{II.36.38} & 82. & 18.52 & {\bf 16.24} & 18.7 & {\bf 15.76} \\
 \text{II.37.1} & 83. & 136.845 & {\bf 46.77} & 92.81 & {\bf 45.08} \\
 \text{II.38.3} & 84. & 122.17 & {\bf 114.56} & 148.47 & {\bf 98.47} \\
 \text{II.38.14} & 85. & {\bf 0.76} & 0.99 & {\bf 0.88} & 1.09 \\
 \text{III.4.32} & 86. & 11.23 & {\bf 7.24} & 9.05 & {\bf 8.04} \\
 \text{III.4.33} & 87. & 41.59 & {\bf 27.14} & 42.31 & {\bf 33.12} \\
 \text{III.7.38} & 88. & {\bf 0.} & 50.43 & 55.54 & {\bf 54.95} \\
 \text{III.8.54} & 89. & {\bf 3.68} & 4.38 & {\bf 4.1} & 4.6 \\
 \text{III.9.52} & 90. & 45.78 & {\bf 38.94} & 46.29 & {\bf 39.17} \\
 \text{III.10.19} & 91. & 69.8 & {\bf 54.84} & 72.05 & {\bf 56.28} \\
 \text{III.12.43} & 92. & 0. & 0. & 4.08 & {\bf 0.} \\
 \text{III.13.18} & 93. & 1152.36 & {\bf 1107.82} & 1206.76 & {\bf 1089.88} \\
 \text{III.14.14} & 94. & 109.04 & {\bf 98.87} & 110.38 & {\bf 97.74} \\
 \text{III.15.12} & 95. & 105.83 & {\bf 105.45} & {\bf 107.98} & 109.96 \\
 \text{III.15.14} & 96. & 8.06 & {\bf 7.1} & 7.84 & {\bf 7.08} \\
 \text{III.15.27} & 97. & {\bf 0.} & 33.66 & {\bf 28.29} & 35.75 \\
 \text{III.17.37} & 98. & 88.19 & {\bf 72.66} & 90. & {\bf 74.54} \\
 \text{III.19.51} & 99. & {\bf 0.61} & 0.72 & {\bf 0.59} & 0.86 \\
 \text{III.21.20} & 100. & {\bf 126.007} & 131.52 & 144.83 & {\bf 132.95} \\
 \hline
\end{tabular}
\end{center}
\end{table}

\end{document}